%% file: main.tex
\documentclass[runningheads]{llncs}

\pdfoutput=1
\usepackage{eccv}

\usepackage{multirow}
\usepackage[normalem]{ulem}
\useunder{\uline}{\ul}{}

\usepackage{eccvabbrv}

\usepackage{graphicx}
\usepackage{booktabs}
\usepackage{wrapfig}

\usepackage{marvosym}

\usepackage[accsupp]{axessibility}  

\usepackage[pagebackref,breaklinks,colorlinks,citecolor=eccvblue]{hyperref}

\usepackage{orcidlink}

\begin{document}

\title{VersaT2I: Improving Text-to-Image Models with Versatile Reward} 


\author{
Jianshu Guo~\inst{1}$^{,*}$ \and
Wenhao Chai~\inst{2}$^{,*\dagger}$ \and
Jie Deng~\inst{1}$^{,*}$ \and
Hsiang-Wei Huang~\inst{2} \and
Tian Ye~\inst{3} \and
Yichen Xu~\inst{4} \and
Jiawei Zhang~\inst{1} \and
Jenq-Neng Hwang~\inst{2}\and
Gaoang Wang~\inst{1}$^{,\text{\Letter}}$
}

\authorrunning{VersaT2I Preprint.}

\institute{
Zhejiang University \and
University of Washington \and
Hong Kong University of Science and Technology (GZ) \and
Fudan University \\
}

\maketitle

\renewcommand{\thefootnote}{\fnsymbol{footnote}}
\footnotetext[1]{Equal contribution, \textsuperscript{$\dagger$}Project Lead, \textsuperscript{\Letter} Corresponding author.}
\renewcommand*{\thefootnote}{\arabic{footnote}}

\input{0_abs}
\input{1_intro}
\input{2_survey}
\input{3_method}
\input{4_exp}
\input{5_ablation}
\input{6_conclusion}

\newpage

\section*{Supplementary Meterials}

\input{7_supp}

\newpage

\bibliographystyle{splncs04}
\bibliography{ref}
\end{document}

%% file: 0_abs.tex
\begin{abstract}
Recent text-to-image (T2I) models have benefited from large-scale and high-quality data, demonstrating impressive performance.
However, these T2I models still struggle to produce images that are aesthetically pleasing, geometrically accurate, faithful to text, and of good low-level quality. We present VersaT2I, a versatile training framework that can boost the performance with multiple rewards of any T2I model. We decompose the quality of the image into several aspects such as aesthetics, text-image alignment, geometry, low-level quality, etc. Then, for every quality aspect, we select high-quality images in this aspect generated by the model as the training set to finetune the T2I model using the Low-Rank Adaptation (LoRA). Furthermore, we introduce a gating function to combine multiple quality aspects, which can avoid conflicts between different quality aspects. Our method is easy to extend and does not require any manual annotation, reinforcement learning, or model architecture changes. Extensive experiments demonstrate that VersaT2I outperforms the baseline methods across various quality criteria. 
\keywords{Generative models \and Text-to-Image Generation \and Improving generative models}
\end{abstract}

%% file: 1_intro.tex
\section{Introduction}

Text-to-image (T2I) generation is currently one of the most prominent research areas in the fields of computer vision and artificial intelligence. It has attracted widespread attention and has profound impacts across industries and other applications, such as computer graphics, art and design, medical imaging, etc. In these works, the emergence of a series of methods represented by diffusion models~\cite{rombach2022high, podell2023sdxl, saharia2022photorealistic} become the dominant method for image~\cite{cao2023difffashion,cao2023image}, video~\cite{chai2023stablevideo}, and 3D~\cite{ouyang2023chasing,deng2023citygen} generation. 

Despite the success of the T2I model, existing T2I models still face challenges in generating images, such as substandard aesthetic quality, limited understanding of the physical properties of realistic scenes~\cite{sarkar2023shadows}, and misalignment between the prompts and the resultant images. Researchers have also proposed various evaluation methods that focus on these different aspects~\cite{hu2023tifa}. Recent work~\cite{black2023training, fan2023dpok} proved that using these evaluation metrics as reward functions can improve the performance of models. These works predominantly employ Reinforcement Learning from Human Feedback (RLHF) based approaches. Unfortunately, these methods require expensive labeled data. Besides, most of these methods sacrifice aesthetic appeal when optimizing for faithfulness. 

\begin{figure}[t]
    \centering
    \includegraphics[width=\linewidth]{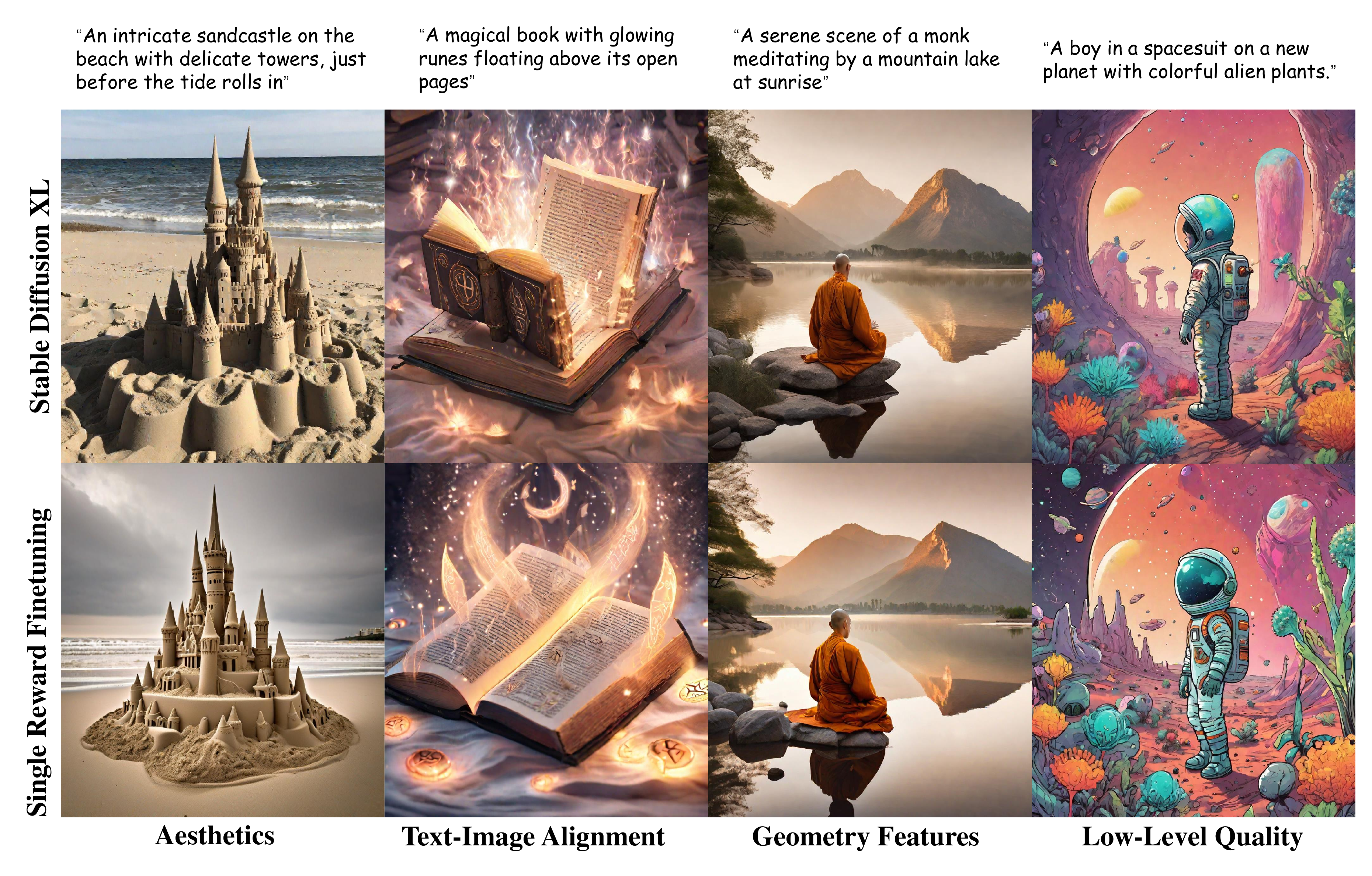}
    \caption{
    We present \textbf{VersaT2I}, a versatile training framework that can boost the performance with multiple rewards of any T2I model. We conduct different reward model under four aspects: aesthetics, text-image alignment, geometry, and low-level quality. Above figure shows the result before and after fine-tuning using SDXL as base model.
    }
    \label{fig:1}
\end{figure}

We propose VersaT2I, a novel framework to address the aforementioned challenges instead of the conventional RLHF paradigm. We decompose the quality of the image into four different aspects: aesthetics, text-image alignment, geometry and low-level quality. Then, we use feedback from models that evaluate these aspects. Specifically, given a T2I model and an evaluation model corresponding to a single quality aspect, we first use a set of prompts to drive the model to generate multiple candidate images per prompt. Then, it automatically evaluates these generated images using the evaluation model. The best generations are collected as a training set to finetune T2I model using parameter-efficient Low-Rank Adaptation (LoRA) finetuning~\cite{hu2021lora}. This method can improve the performance of the model in each specific aspect. 

It's a natural extension to combine multiple reward signals to improve the overall quality of the image. However, exists works using the reinforcement learning paradigm to combine multiple reward signals has the following disadvantages: First, reinforcement learning requires a tedious process, which is aggravated by multiple rewards, and is both time-consuming and resource-intensive. Secondly, it is also not scalable as the number of rewards increases. Furthermore, there are conflicts between multiple metrics. Optimizing one metric may cause the performance degradation of another metric. Therefore, it is not easy to find a balance point.

We expand our method beyond a single aspect to include multiple aspects. We develop a method called Mixture of LoRA (MoL). This method is inspired by the Mixture of Experts (MoE) and involves combining the LoRA models that focus on different aspects to enhance the model's overall performance. Concretely, at each LoRA layer of the model, we replace each LoRA layer with multiple parallel LoRA and connect them with routers. We then propose using localized balancing constraints to balance the importance of all experts, which prevents only a few experts from being valued by the routers. Experiment results show that this method can improve the model's performance in various quality aspects.

Our contributions and advantages are as follows: 
\begin{enumerate}
\item We propose VersaT2I: a self-training and model-agnostic framework that can combine any number of evaluation models and does not require cumbersome RL optimization processes.  
\item We design a self-training method that uses the data generated by the model for training without requiring additional training data. Our method does not require expensive human annotation.
\item We introduce the Mixture of LoRA method, which effectively combines four different quality aspects trained by individual evaluation models to enhance the overall quality of generated images. 
\end{enumerate}

%% file: 2_survey.tex
\section{Related Work}

\subsection{T2I Models}
The goal of T2I models is to create an image given an input text prompt. Several T2I generative models have been proposed and have demonstrated promising results~\cite{rombach2022high, podell2023sdxl}. Among them, Stable Diffusion models \cite{podell2023sdxl} show impressive generation performance in T2I generation. Despite substantial progress, the images generated by those models still exhibit quality issues, such as lousy cropping or misalignment with the input texts.

\subsection{Evaluating T2I Models}
Early T2I models are only trained on small-scale dataset such as CUB birds~\cite{wah2011caltech}, Oxford flowers~\cite{nilsback2008automated}, MS-COCO~\cite{lin2014microsoft}, and ImageNet~\cite{deng2009imagenet}. Several metrics like Fréchet Inception Distance~(FID)~\cite{heusel2017gans}, Inception Score~(IS)~\cite{salimans2016improved}, Learned Perceptual Image Patch Similarity~(LPIPS)~\cite{zhang2018unreasonable}, and CLIPScore~\cite{hessel2021clipscore} are commonly used for T2I model evaluation. Recently, as the capabilities of T2I models gradually become more powerful, some methods based on human evaluation and multimodal large language model evaluation have developed. 
HPSv2~\cite{wu2023human} and ImageReward~\cite{xu2023imagereward} systematically establish human preference annotation for T2I models.
TIFA~\cite{hu2023tifa}, VIEScore~\cite{ku2023viescore}, LLMscore~\cite{lu2023llmscore}, DSG~\cite{cho2023davidsonian} further leverage multimodal large language model, using the Visual Question Answering (VQA) task format as quantity evaluation.
Other works focus on object attribute and relationship evaluation, like T2I-CompBench~\cite{huang2023t2i} and CLIP-R-Precision~\cite{park2021benchmark}.
Moreover, HEIM~\cite{lee2023holistic} identifies 12 evaluation aspects, including text-image alignment, image quality, aesthetics, originality, reasoning, knowledge, bias, toxicity, fairness, robustness, multilingualism, and efficiency. In VersaT2I, we use a set of evaluation models in terms of four different essential aspects: aesthetics, geometry, text-faithful, and low-level vision.

\subsection{Improving T2I Models}
Many works are trying to improve T2I models with multiple aspects. ControlNet~\cite{zhang2023adding} adds various conditioning controls, e.g., edges, depth, segmentation, human pose, etc., by hypernetwork structure. DreamBooth~\cite{ruiz2023dreambooth} presents an approach for synthesizing novel renditions of a subject using a few images of the subject.
Other works focus on customization of style~\cite{sohn2023styledrop} or object relation~\cite{huang2023reversion}.
Beyond those few-shot settings, many works also try to use reinforcement Learning to improve T2I models. 
Rich Automatic Human Feedback~(RAHF)~\cite{liang2023rich} learn to predict rich human
annotations on generated images and their associated text prompt.
Reward Feedback Learning~(ReFL)~\cite{xu2023imagereward} leverages the feedback from ImageReward~\cite{xu2023imagereward} to optimize diffusion models at a random latter denoising step directly.
Diffusion Policy Optimization with KL regularization~(DPOK)~\cite{fan2023dpok} utilizes KL regularization w.r.t. the pre-trained text-to-image model as an implicit reward to stabilize RL fine-tuning. Parrot~\cite{lee2024parrot} presents a Pareto-optimal multi-reward reinforcement learning framework.
Besides, Diffusion-DPO~\cite{wallace2023diffusion} uses Direct Preference Optimization (DPO) for aligning diffusion models to human preferences by directly optimizing the model on user feedback data.
DreamSync~\cite{sun2023dreamsync} leverages training but does not involve reinforcement learning.
Compared to the above method, VersaT2I uses more versatile reward models and avoids unstable training based on RL. At the same time, by adopting a reasonable merge strategy for LoRAs trained with different reward models, we also prevent the degradation phenomenon caused by conflicts among multiple reward models.

%% file: 3_method.tex
\section{Methods}

\begin{figure}[t]
    \centering
    \includegraphics[width=\linewidth]{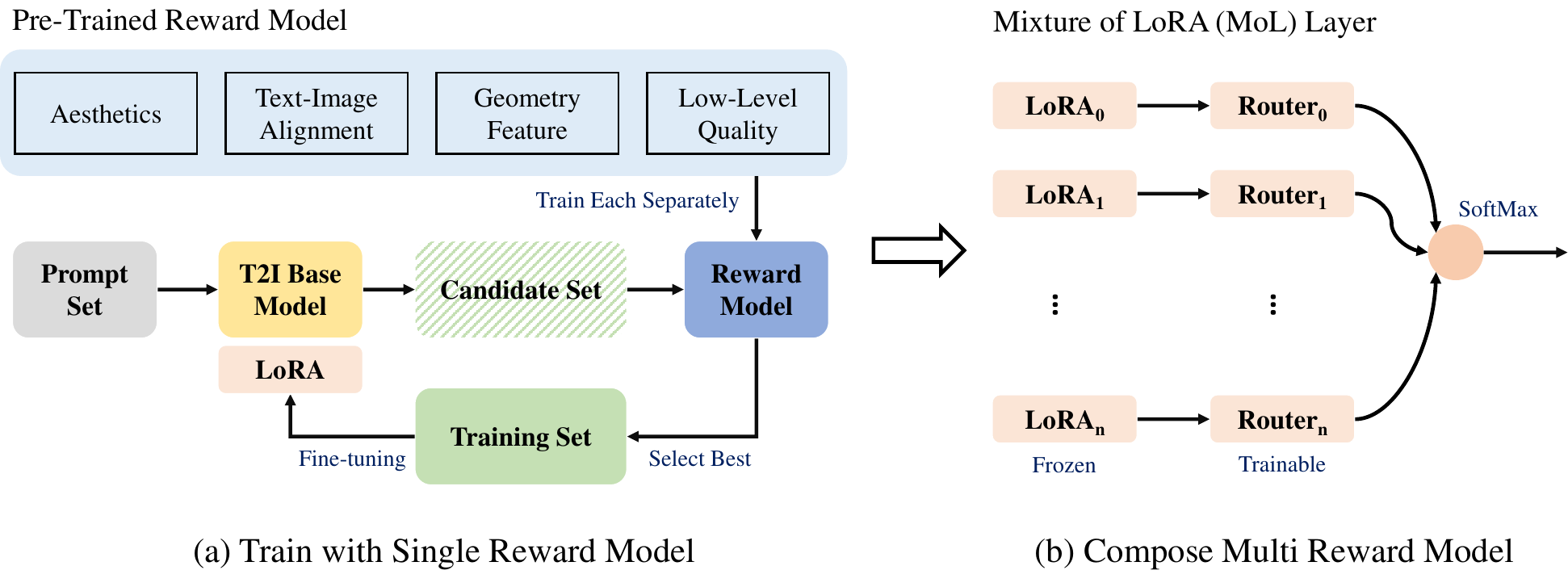}
    \caption{Overview of our proposed VersaT2I framework under two-stage training. Firstly, given a prompt, a text-to-image generative model generates a batch of images as a candidate set, which then is evaluated by a pre-trained reward model. After that, we fine-tune the T2I model using LoRA with samples of the best score among the candidate sets as the training set. After obtaining LoRA for each reward model, we further compose multiple LoRAs in a Mixture of LoRA layer design to achieve versatile improvement.}
    \label{fig:framework}
\end{figure}

\subsection{Preliminary}
\textbf{Diffusion Models}~\cite{ho2020denoising, rombach2022high} are state-of-the-art generative models known for their high-quality, photorealistic image synthesis. It operates on the principle of simulating a stochastic process where Gaussian noise is gradually added to an image at each step, and the model is trained to reverse this process, removing the noise and reconstructing the original image. The reverse process is typically learnt using an UNet with text conditioning support enabling text-to-image generation at the time of inference. Specifically, given an initial noise map $\epsilon \sim \mathcal{N}(0, I)$ and a text-image pair $(c, x)$, they are trained using a squared error loss to denoise a variably-noised image as follows: 
\begin{equation}
    \mathbb{E}_{x,c,\epsilon,t}[w_t||\hat{x}_\theta(\alpha_t x + \sigma_t \epsilon, c) - x ||_2^2]
    \label{equation 1}
\end{equation}
where x is the ground-truth image, c is a text prompt, and $\alpha_t, \sigma_t, w_t$ are terms that control the noise schedule and sample quality.

\vspace{5pt}
\noindent\textbf{LoRA~(Low-Rank Adaptation) Fine-tuning} is a method for efficient fine-tuning of Large Language and Vision Models. Instead of fine-tuning all weights, LoRA decomposing weight residuals $\Delta W\in \mathbb{R}^{m\times n}$ into two low rank matrix $A\in \mathbb{R}^{m\times r}$ and $B\in \mathbb{R}^{r\times n}$ for efficient parameterization with $\Delta W=BA$:
\begin{equation}
    W+\Delta W = W + BA
\end{equation}
Here $r$ represents the intrinsic rank of $\Delta W$ with $r << min(m, n)$. During training update the values of $A$ and $B$ instead of $W$, thus greatly reducing the number of training parameters. During inference, the W is frozen:
\begin{equation}
    h=Wx+\alpha BAx
    \label{equation3}
\end{equation}
where $\alpha$ represents a scaling factor that adjusts the magnitude of the changes on the original $W$ made by LoRA modules.

\vspace{5pt}
\noindent\textbf{The Mixture of Experts} scales up model parameters without correspondingly increasing computational efforts. For transformers-based large language models, MoE supplants the conventional feed-forward neural network layer in each transformer block with an MoE layer~\cite{shazeer2017outrageously}. The MoE layer comprises $N$ parametrically identical and independent feed-forward neural networks as the experts, coupled with a gating function $G(\cdot)$ as the router. The router is used to model the probability distribution that governs the weights of outputs from these expert networks. Formally, for the output $h$ of the attention layer in any given block, the output $y$ of the MoE layer can be formulated as follows:
\begin{equation}
    y=\sum\limits_{i=1}^N G(h)_i E_i(h)
\end{equation}
where $E_i(h)$ and $G(h)_i$ denote the output and the corresponding weight of $i$-th expert in the MoE layer, respectively. The router $G(\cdot)$ can be written as follows: 
\begin{equation}
    G(\cdot) = Softmax(hW_g)
\end{equation}
where $W_g$ is the trainable weight matrix for router $G(\cdot)$.

\subsection{VersaT2I Overview}
In~\cref{fig:framework} we show the framework of VersaT2I, which consists of the training set generation stage, the single reward training stage, and the multi-reward combination stage. We first decompose image quality into four aspects: aesthetics, low-level quality, text-image alignment, and projective geometry. The low-level quality refers to finer-grained abilities to identify the low-level visual attributes such as brightness, noise, and blur.  Projective geometry comes from the insight of~\cite{sarkar2023shadows}. Natural images have the correct perspective geometry features. The image generated by the generative model may have a different perspective field than the real image, and the object shadow relationship in the generated image may differ from the object shadow relationship in the real image. Generative models require an understanding of projective geometry to generate realistic images. Therefore, projective geometry can also be used to indicate the quality of generated images. 

In the training set generation stage, we prompt the T2I model to generate a set of $N \times K$ candidate images, where $N$ refers to the number of prompts and $K$ refers to the number of generated images per prompt. Then, we use the corresponding evaluation model for each quality aspect to score the candidate image set and filter out the top $10\% - 20\%$ as the training set. Note that now we have a training set for each aspect, and the training sets for different aspects are different. For example, there may be images with high aesthetic scores but low text-image alignment scores. Such images will be selected into the aesthetic training set but will not appear in the training set corresponding to the text-image alignment. 

Given a training set in the single reward training stage, we fine-tune the generative model on this set using LoRA. For every aspect of quality, we have a corresponding LoRA expert. We argue that by merging LoRAs in some way, we can achieve the effect of improving all qualities at the same time. However, merging multiple LoRAs directly does not improve performance, just as~\cref{table2} shows. In the multi-reward combination stage, we design a method named Mixture of LoRA~(MoL), which uses a gating function to determine the weight of different LoRAs automatically. 

\subsection{Train Single Reward}

Previous work mainly used RL-based methods to improve the performance of the model. Our approach uses a simple yet powerful approach. Generating samples from the model and evaluating these samples with a scoring mechanism. Then, we select a batch of samples with the highest scores as the training set. The original model undergoes supervised fine-tuning on the training dataset. 

\vspace{5pt}
\noindent\textbf{Build Prompt Set.} To ensure the training set achieves adequate generalization, we need comprehensive prompts so that the model can generate sufficient diverse images. We make full use of the in-context learning capabilities of Large Language Models (LLMs) and use GPT-4 to generate diverse prompts. We add some instructions, such as specifying the category of elements in prompts, no repetition, etc. We show the specific process in the appendix. We denote the prompt set as $\mathbb{P}=\{ P_i \}_i^N$, where $N$ represents the number of prompts.

\vspace{5pt}
\noindent\textbf{Build Candidate Set.} Given the prompt set $\mathbb{P}=\{ P_i \}_i^N$, we prompt the T2I model to randomly generate candidate images. Because the images are randomly generated, the images generated by the model vary from good to bad. In theory, higher-quality images in the training set are preferable. The more images generated by the model, the higher the probability of high-quality images appearing. We generate $K$ samples with each prompt $P_i$. Prior work~\cite{karthik2023if} considers 5-10 samples per prompt enough to produce a good image. Finally, the candidate set has $M = N \times K$ images.

\vspace{5pt}
\noindent\textbf{Build Training Set.} We use an evaluation model corresponding to a single quality aspect to score each image in the candidate set. Suppose there exists an evaluation model $E$, which produces scores for each prompt $P_i$ denoted as $\mathbb{S}_E(P_i) = \{ S_E(I_{i,k}) \}_{k=1}^K$ for the candidate set. We set a threshold $\theta$ to filter out the highest-scoring portion of images.  Formally, for a text prompt $P_i$ and its corresponding synthetic images $\{ I_{i,k}\} _{k=1}^K$, we select samples that pass the threshold $\theta$: $\mathbb{D}(P_i) = \{(P_i, I_{i,k}):S_E(I_{i,k})>=\theta \}$. However, there is an imbalance issue for such sampling. Some simpler prompts may generate more high-quality images, while difficult prompts may even generate no high-quality images. This results in an imbalanced distribution where easy prompts
have more samples, which could cause degradation of image quality. To avoid it, we only select one representative image $\hat{I}_i$ having the highest score for each $P_i$, \textit{i.e.}, $\hat{I}_i=\mathop{\arg\max}\limits_{I \in \{I_{i,k}\}_{k=1}^K}\mathbb{S}_E(P_i)$.
We apply this procedure to all text prompts to build the training set $\{(P_i,\hat{I}_i)\}_{i=1}^N$ for each quality aspect. 

\vspace{5pt}
\noindent\textbf{LoRA Fine-tuning.} After obtaining a training set for each quality aspect, we fine-tune our generative using LoRA on this set. Given a generative model $G$ and the pre-trained weight matrix $W_0 \in \mathbb{R}^{m \times n}$, instead of directly fine-tuning $W_0$, we perform low-rank decomposition of the increment matrix $\Delta W$: 
\begin{equation}
    W = W_0 + \Delta W
\end{equation}
\begin{equation}
    \Delta W = AB,    A \in \mathbb{R}^{m \times r}, B \in \mathbb{R}^{r \times n}
\end{equation}
where $W_0$ remains unchanged and only $A$ and $B$ are updated. We apply LoRA for the attention layers of the generative model $G$. We fine-tune $G$ using the training objective~\cref{equation 1} on the training set.

\subsection{Compose Multi Reward}
Each LoRA can improve the performance of a quality aspect. A natural extension is combining multiple LoRAs to enhance the performance of multiple model aspects simultaneously. One of the most straightforward approaches is directly combining all the LoRAs, as proposed in previous works\cite{zhang2024composing}. However, we find this approach to be limited. The combination of multiple LoRA modules leads to a degradation in generation capability compared to individual LoRA modules. We argue the issue with the combination method is the uniform weighting of one LoRA across all layers. The different LoRA layers contribute to the final results in varying degrees. Unified coefficients will cause the effects of different LoRAs to conflict with each other.

We introduce the Mixture of LoRAs to integrate multiple LoRAs corresponding to different quality aspects effectively. It draws inspiration from gating mechanisms to model the weight distribution of the set of frozen LoRAs at each layer, aiming to find the most suitable combination for these LoRAs. As shown in~\cref{fig:framework}, we replace each LoRA layer with an MoE-like architecture. Specifically, we freeze the backbone model's parameters and each LoRA module's parameters, retaining its original strong generative capabilities. Formally, the forward process of the Mixture of LoRA layer can be expressed as: 
\begin{equation}                        
    o = W_0x + \sum\limits_{i=1}^L G(x)_i \Delta W_ix
\end{equation}
where $G(.)=Softmax(xW_g)$ represents the router in the MoL layer; $\Delta W_i$ represents the LoRA matrix $B_i A_i$; the $W_g$ is the trainable parameter matrix of the route function; and $L$ is the number of LoRAs.

\vspace{5pt}
\noindent\textbf{Training Objective.} We adapt the diffusion loss $\mathcal{L}_0$~from~\cref{equation 1} to train the route function. Besides, during the competition between different experts, the gating function tends to converge to a state where it always produces large weights for an early-stage well-performing expert~\cite{shazeer2017outrageously}, leading to only one of the LoRA experts having a significant impact in the end. To alleviate this imbalance across multiple LoRA experts, we propose a gating balancing loss:
\begin{equation}
     \mathcal{L}_1 = -log\prod\limits_{i=0}^L{p^{(i)}}
\end{equation}
where
\begin{equation}
     p^{(i)} = \frac{1}{L}\sum\limits_{j=1}^L Softmax(xW_g)_j
\end{equation}
The overall training objective is the weighted sum of the two losses: 
\begin{equation}
    \mathcal{L} = \mathcal{L}_0 + \alpha \mathcal{L}_1
\end{equation}
where $\alpha$ is a coefficient for weight balancing, the gating balanced loss encourages a balanced route function because it is minimized when the dispatching is balanced. 

%% file: 4_exp.tex
\section{Experiments}
\subsection{Implementation Details}
\textbf{Dataset}
To obtain prompts, we utilize the in-context learning capability of Large Language Models. We prompt GPT-4 to generate 12k non-repetition prompts. At the same time, we include additional instructions that specify the category involved in the prompt, such as shape, counting, color, location, animal/human~\cite{hu2023tifa}.

\vspace{5pt}
\noindent\textbf{Base Model.} We evaluate every single LoRA model and the Mixture of LoRAs on Stable Diffusion v2.1 and Stable Diffusion XL\cite{podell2023sdxl}. SDxl is the current state-of-the-art open-sourced T2I model. As a result, we mainly demonstrate the effect on SDxl. For each prompt, we generate eight images per prompt.

\vspace{5pt}
\noindent\textbf{Training Details.} For LoRA finetuning, all four models use the same set of parameters. We conduct experiments on a machine equipped with 4 NVIDIA GTX 4090 GPUs. DPM++ with 30 steps is used for sampling, and the classifier-free guidance weight is set to 7.0 with the resolution $1024\times1024$ for SDXL, $768\times768$ for SD v2.1. For LoRA fine-tuning, we employ the Adam optimizer and the Cosine Scheduler with an initial learning rate of 1e-4. Instead of updating all layers, we specifically update the cross-attention and self-attention layers in the denoising U-Net. For single LoRA inference, we set the LoRA $\alpha$ in~\cref{equation3} to $0.5$. For the Mixture of LoRAs fine-tuning, we freeze both the base model and all LoRA modules and only optimize the gating function. We use Adam optimizer with an initial learning rate of 1e-5.

\subsection{Reward Model Selection}

\noindent\textbf{Aesthetic Evaluation.} For aesthetic assessment model, we use the Q-Align~\cite{wu2023q}, which is a state-of-the-art image aesthetic assessment (IAA) model. Q-Align teaches Large Multi-Modality Models (LMMs) for visual rating aligned with human opinions through a structured methodology. During training, the process simulates training human annotators by converting Mean Opinion Score (MOS) values to five text-defined rating levels. These levels are then used to conduct visual instruction tuning on LMMs. In the inference stage, the strategy simulates collecting MOS from human ratings by extracting log probabilities on different rating levels and using softmax pooling to obtain close-set probabilities for each level. The final LMM-predicted score is derived from a weighted average of these probabilities. This approach has proven to be more effective than using direct scores as learning targets, leading to significant improvements in visual scoring tasks such as image quality assessment (IQA), image aesthetic assessment (IAA), and video quality assessment (VQA). We use Q-Align to measure the aesthetic score. Empirically, we keep the text-image pairs whose aesthetic scores are greater than $\theta_{aes} = 4.9$. If multiple generated images pass the threshold, we keep the one with the highest aesthetic score. Starting from 12,177 prompts, we have 1,900 images exceeding this threshold.

\vspace{5pt}
\noindent\textbf{Geometrical Feature Evaluation.} For the geometrical evaluation model, we use a model that detects geometric features proposed in~\cite{sarkar2023shadows}. The model utilizes Perspective Fields\cite{jin2023perspective}, vector fields that encode the spatial orientation of pixels in relation to vanishing points and the horizon. These dense fields could be instrumental in assessing the projective geometry of images.  We use a pre-trained model to generate Perspective Fields from single images, which serve as a basis for understanding the scene’s geometric structure. Then, we train a ResNet50 classifier on these fields to differentiate between real and generated images. The classifier evaluates the consistency of Perspective Fields with projective geometry principles, scoring images on their geometric plausibility. After filtering, we have 5,986 images as the training set. 

\vspace{5pt}
\noindent\textbf{Text-faithful Evaluation.} For text-image alignment, we use the VQA model proposed in \cite{hu2023tifa}, a simple tool to evaluate fine-grained text-image alignment by asking and answering questions about it, utilizing the power of Large Language Models. To evaluate the faithfulness of the generated images to the textual input, TIFA uses VQA models to check whether, given a generated image, questions about its content are answered correctly. We use the LLaMA 2 fine-tuned in TIFA for question generation and use mPLUG-large~\cite{li2022mplug}  as the VQA model to measure the faithfulness of generated images to textual input. Finally, we set the $\theta_{text} = 1.0$, which means that the text and image are completely aligned. Despite these challenges, a total of 5,834 images were selected. 

\begin{table}[t]
\caption{\textbf{Single Reward Model Result.} Each column in the table represents the results from training with \textbf{a single reward}. All models are sampled with the same set of eight seeds. Best scores under each backbone T2I model are highlighted in \textbf{bold}. Our approach exhibits the best performance in most situations. }
\label{table1}
\centering
\resizebox{\linewidth}{!}{
\begin{tabular}{@{}l|l|cccc@{}}
\toprule
Base                    & ~Method~ & ~Aesthetically($\uparrow$) & Text-faithful($\uparrow$) & Geometrically($\uparrow$) & Low-level quality($\uparrow$) \\ \midrule
\multirow{3}{*}{SD v2.1~} & ~None   & 3.89          & 81.27          & 0.26         & 0.623             \\ 
                         & ~DDPO   & 3.87          & 81.45         & 0.26         & 0.624             \\ 
                         & ~\textbf{Ours}   & \textbf{3.90} & \textbf{81.87} & \textbf{0.28}& \textbf{0.624}             \\ \midrule
\multirow{3}{*}{SDXL}    & ~None   & 4.26          & 83.32          & 0.32          & 0.660             \\ 
                         & ~DDPO   & 4.29          & 83.91          & 0.31         & \textbf{0.664}             \\ 
                         & ~\textbf{Ours}   & \textbf{4.30} & \textbf{84.29} & \textbf{0.35}& 0.663             \\ \bottomrule
\end{tabular}
}
\label{table12}
\vspace{-12pt}
\end{table}

\vspace{5pt}
\noindent\textbf{Low-level Quality Evaluation.} For the low-level evaluation model, we use the Q-Instruct~\cite{wu2023q-instruct}, a Multi-modality Foundation Model with instruction tuning on a large-scale low-level visual instruction tuning dataset. The Q-Instruct model aims to improve various low-level visual abilities of multi-modality foundation models by providing diverse instruction types, including visual question-answering and extended conversation subsets. These additional instruction types allow the models to respond to various human queries and enhance their low-level perception abilities. It is a valuable resource for training multi-modality foundation models to excel in low-level visual tasks~\cite{ye2021perceiving,ye2022towards,liu2023nighthazeformer,chen2023dehrformer,chen2022msp,chen2022snowformer,chen2023sparse,ye2022underwater,jiang2023five,zou2022self,liu2022nighttime,ye2023adverse,chen2023uncertainty,zou2023vqcnir} and improve their overall performance in understanding and processing visual information. We set the threshold $\theta_{low-level}$ to 0.8, where 878 images were selected. 

\begin{table}[t]
\centering
\caption{Our MoL merging method results compared with directly merging. We find that directly merging's perfermance is even slightly worse than the original model, while our merging strategy can maintain the performance from each reward model.}
\resizebox{\linewidth}{!}{
\begin{tabular}{l|l|cccc}
\toprule
Base                   & ~Method           & ~Aesthetically($\uparrow$) & Text-faithful($\uparrow$) & Geometrically($\uparrow$) & Low-level quality($\uparrow$) \\ \midrule
\multirow{3}{*}{SD v2.1~} & ~None             & 3.89          & 81.27         & \textbf{0.26}          & 0.623             \\ 
                         & ~Directly merging~ & 3.89          & 81.31         & 0.24          & 0.621             \\ 
                         & ~MoL              & \textbf{3.91}          & \textbf{81.58}         & \textbf{0.26}          & \textbf{0.625}             \\ \midrule
\multirow{3}{*}{SDXL}    & ~None             & 4.26          & 83.32         & 0.32          & 0.660             \\
                         & ~Directly merging~ & 4.27          & 83.91         & 0.33          & 0.660             \\ 
                         & ~MoL              & \textbf{4.29}          & \textbf{84.18}         & \textbf{0.34}          & \textbf{0.661}             \\ \bottomrule
\end{tabular}
}
\label{table3}
\end{table}

\begin{table}[t]
\centering
\caption{\textbf{Ablation study} on the effectiveness of gating balancing loss. Each value in the table corresponds to the coefficient of the respective gating function. With the balance loss, we observe that the coefficients across different quality aspects are more evenly distributed.}
\resizebox{0.95\linewidth}{!}{
\begin{tabular}{l|cccc}
\toprule
Setting~          & ~Aesthetically & Text-faithful & Geometrically & Low-level quality \\ \midrule
  w/o. Balance Loss~     & 0.713          & 0.068         & 0.096          & 0.123             \\ 

w. Balance Loss~ & 0.428          & 0.169         & 0.163          & 0.240             \\ \bottomrule
\end{tabular}
}
\label{tab:table4}
\end{table}

\begin{figure}
    \centering
    \includegraphics[width=\linewidth]{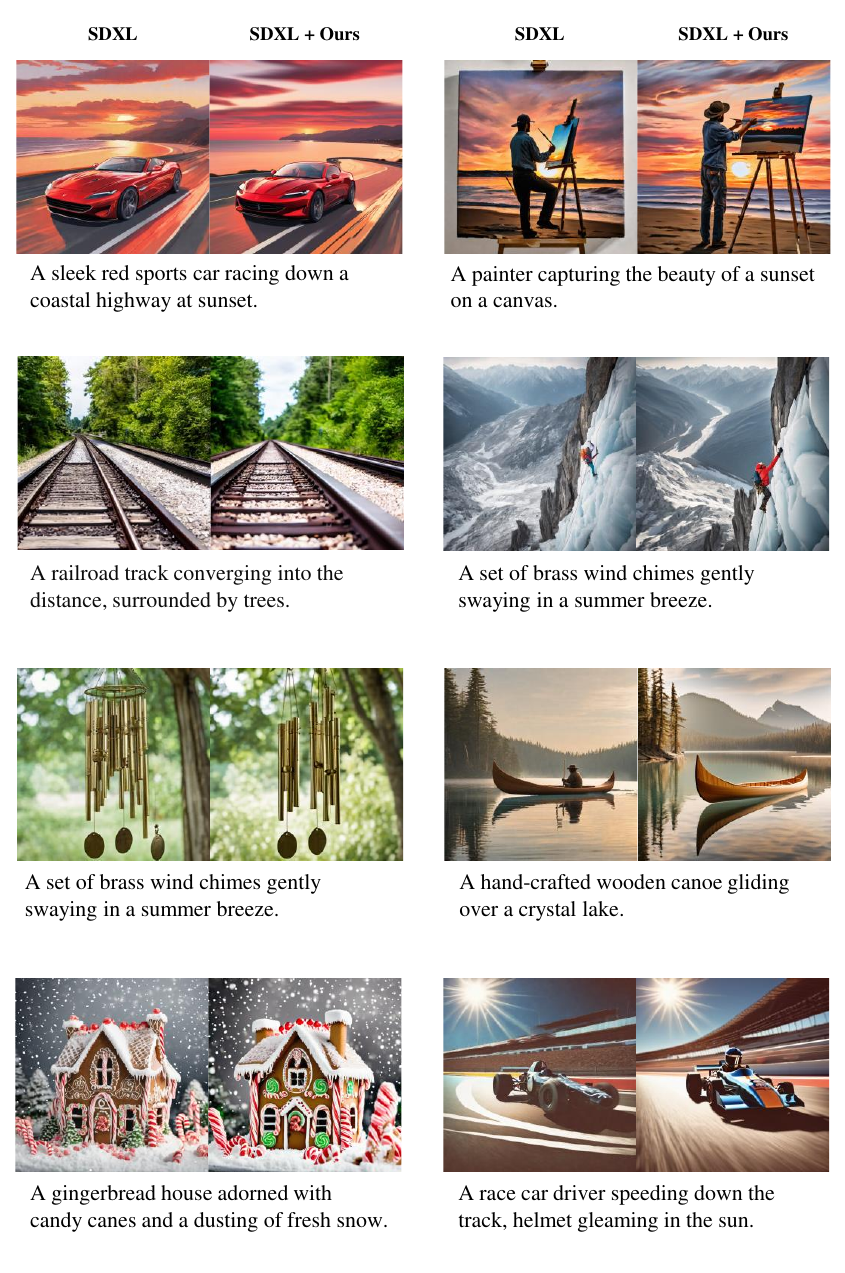}
    \caption{\textbf{Qualitative results} of proposed VersaT2I improving aesthetics, text-image alignment, geometry, low-level quality after training. All the image resolution is $1024 \times 1024$ and generated with same noise and seed for fair comparison.}
    \label{fig:sdxl_vs_ours}
\end{figure}

\subsection{Single Reward Model Results}
In~\cref{table12} we compare single reward LoRA to some state-of-the-art approaches. The single reward LoRA improves all of the evaluation benchmarks of both SD v2.1 and SDXL. For SDXL, the aesthetic score increased by 0.01, the Geometrical score increased by 0.03, the Text-faithful (TIFA benchmark) score increased by 0.97, and the low-level score (Q-instruct benchmark) increased by 0.003. We report more results in the Appendix.

\subsection{The Mixture of LoRA Results}
\textbf{Baseline.} We compare our method with directly merging LoRA by fixed coefficient\cite{zhang2024composing}, just as:
\begin{equation}
    W = W_0 + \alpha\sum\limits_{i=1}^N \omega_i\Delta W_i
\end{equation}
where $W$ indicates the final combination weight and $\omega_i$ denotes the weights of different LoRAs. We set $\sum\limits_{i=1}^N \omega_i =1$ to prevent any adverse impact on the embedding of the original model. We set the LoRA $\alpha$ value to be the same (i.e. avarage) during the single LoRA inference.

\vspace{5pt}
\noindent\textbf{The Baseline Results.} When directly merging multiple LoRAs, we found that the effect decayed back to the original model. As the \cref{table12} shows, we directly merged two loras to verify the effect of the baseline. The results show that directly merging will make LoRA's performance almost disappear. 

\vspace{5pt}
\noindent\textbf{The Mixture of LoRAs Results.} As shown in~\cref{table3}, our method consistently outperforms the baseline method no matter which LoRA. These results indicate that our method demonstrates robust capabilities, and it can alleviate conflicts between different LoRAs.

\subsection{Qualitative Results}

We show qualitative comparisons of SDXL and our model VersaT2I in~\cref{fig:sdxl_vs_ours}, we also show how each single reward model boost the performance shows in~\cref{fig:1} as well. Obviously, the aesthetic score is improved in the case of "car" and "candy house", the text faithfulness is improved in the case of "canvas" and "wooden canoe", and the geometrical feature is improved in the case of "trail", etc.

%% file: 5_ablation.tex
\section{Ablation Study}

\begin{figure}[t]
    \centering
    \includegraphics[width=\linewidth]{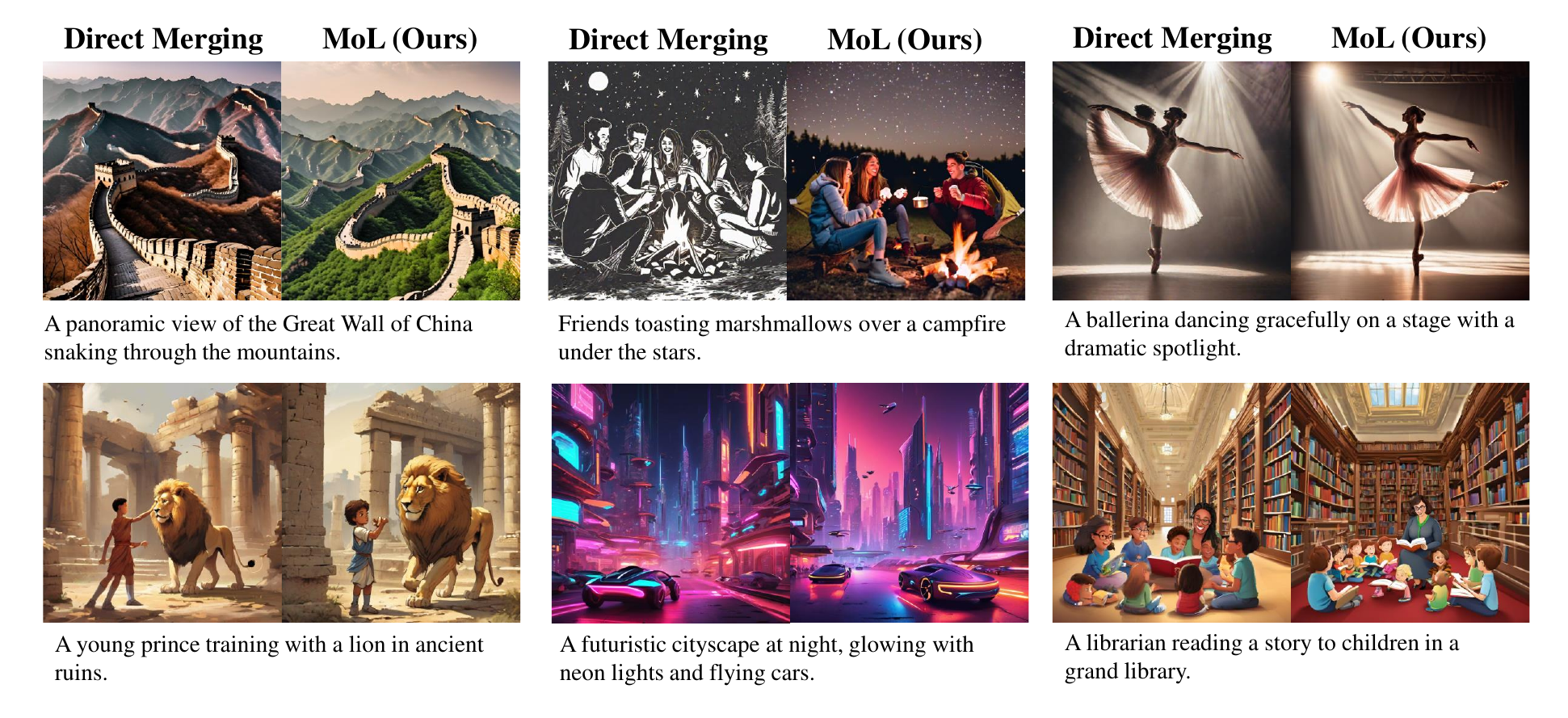}
    \caption{
    \textbf{Ablation study} on Mixture-of-LoRA design qualitatively. Our proposed method acheives better performance compared to naively average the LoRA weights from different reward models. All the image resolution is $1024 \times 1024$ and generated with same noise and seed for fair comparison.
    }
    \label{fig:ablation_mol}
\end{figure}

\noindent\textbf{The effectiveness of gating balancing loss.} The model exhibits a more uniform average weight distribution across all gate functions of each quality aspect. For instance, as depicted in~\cref{tab:table4}.

\vspace{6pt}
\noindent\textbf{Qualitative results comparison} between directly LoRA merging and the proposed MoL as shown in~\cref{fig:ablation_mol}. We found that naively merge all the LoRA will bring degradation performance in some aspect, but our MoL design successfully addresses this issue.

%% file: 6_conclusion.tex
\section{Limitations and Social Impact}

\noindent\textbf{Limitations.} The VersaT2I framework enhances the potential for creating manipulated content and deepfakes, raising concerns about misinformation, privacy violations, and the reinforcement of biases. These issues underline the importance of addressing the social and ethical impacts of generative models, emphasizing the need for future research in this area to mitigate potential harms associated with their use. For future work, one direction is to use more fine-grained annotations with more reward models.

\vspace{6pt}
\noindent\textbf{Social Impact.} Generative models for media bring both benefits and challenges. They foster creativity and make technology more accessible, yet pose risks by facilitating the creation of manipulated content, spreading misinformation, and exacerbating biases, particularly affecting women with deep fakes. Concerns also include the potential exposure of sensitive training data collected without consent. Despite diffusion models potentially offering better data representation, the impact of combining adversarial training with likelihood-based objectives on data distortion remains a crucial research area. Ethical considerations of these models are significant and require thorough exploration.

\section{Conclusion}

This paper introduces VersaT2I, a novel framework designed to enhance Text-to-Image (T2I) models by addressing challenges such as suboptimal aesthetic quality, lack of understanding of physical scene properties, and misalignment between prompts and generated images. Unlike traditional methods that rely on Reinforcement Learning from Human Feedback (RLHF) and require costly human-annotated data, VersaT2I employs a self-training approach that decomposes image quality into four aspects: aesthetics, text-image alignment, geometry, and low-level quality. It leverages evaluation models for each aspect to select and fine-tune generated images using a parameter-efficient LoRA fine-tuning method. The paper also introduces a Mixture of LoRA (MoL) method inspired by the Mixture of Experts (MoE) model, which combines different LoRA models to enhance overall image quality. This approach avoids the downsides of traditional reinforcement learning by being less resource-intensive and avoiding conflicts between multiple reward signals. The experiments demonstrate that VersaT2I outperforms existing methods across several quality aspects of T2I models, offering a scalable, efficient, and versatile framework for improving T2I generation without requiring extensive human-labeled datasets.

%% file: 7_supp.tex
\renewcommand{\thesection}{\Alph{section}}
\setcounter{section}{0}



\section{Prompts Generation Details}

We use GPT-4~\cite{achiam2023gpt} to generate prompts. Here are the detailed instructions we use:
\textbf{Instruction for prompt generation.} \textsf{Please generate similar examples to the provided ones and  follow these guidelines: \\
1. Your generation will be served as prompts for Text-to-Image models. So your prompt should be as
visual as possible.\\
2. The generated examples should be as creative as possible.\\
3. The generated examples should not have repetition.\\
4. The generated examples should be as diverse as possible.\\
5. The generated examples should not be more than 20 words.\\
Here are five examples, please generate similar examples: \\
example1; example2; example; example; example5.}

\noindent We change the seed and repeat the instruction, generating 12k prompts. 

\section{Results on HPSv2 Benchmark}
Human Preference Score v2 (HPSv2) ~\cite{wu2023human} is a solid benchmark for evaluating human preferences of Text-to-Image synthesis. It is trained on human preference dataset comprising a large of human preference choices on pairs of images. We test our model on HPSv2 benchmark to demonstrate that our model aligns with human values and judgments.  As shown in \cref{tab:hps}, we find that our VersaT2I can improve human preference score on both SD 2.1 and SDXL, thus demonstrating that VersaT2I is aligned with human preference.

\begin{table}[ht]
\centering
\caption{HPSv2 benchmark.}
\resizebox{0.85\linewidth}{!}{
\begin{tabular}{l|cccc}
\toprule
Setting~      & SD 2.1~ & SD 2.1+VersaT2I~ & SDXL~ & SDXL+VersaT2I~\\
\midrule
HPSv2~($\uparrow$)~     & 26.96~   & 28.32~   & 32.03~   & 32.96~     \\ 
\bottomrule
\end{tabular}
}
\label{tab:hps}
\end{table}

\clearpage
\section{More Visual Examples}

\begin{figure}
    \centering
    \includegraphics[width=\linewidth]{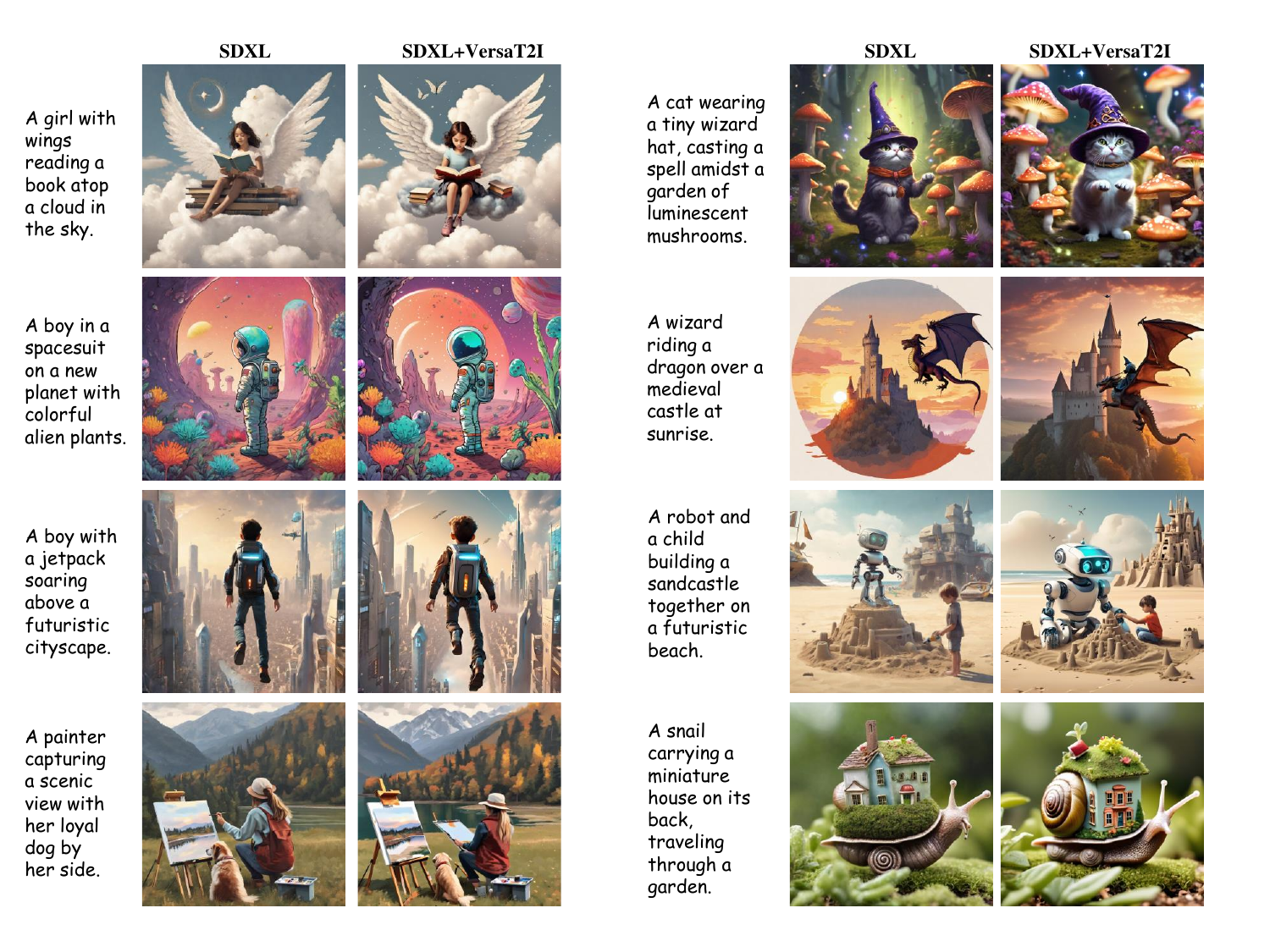}
    \caption{More examples generated by VersaT2I. Given the text prompt and fixed seed, we generate images with SDXL\cite{podell2023sdxl} and VersaT2I. Images generated by VersaT2I exhibit high quality.}
    \label{fig:s1}
\end{figure}

\begin{figure}
    \centering
    \includegraphics[width=\linewidth]{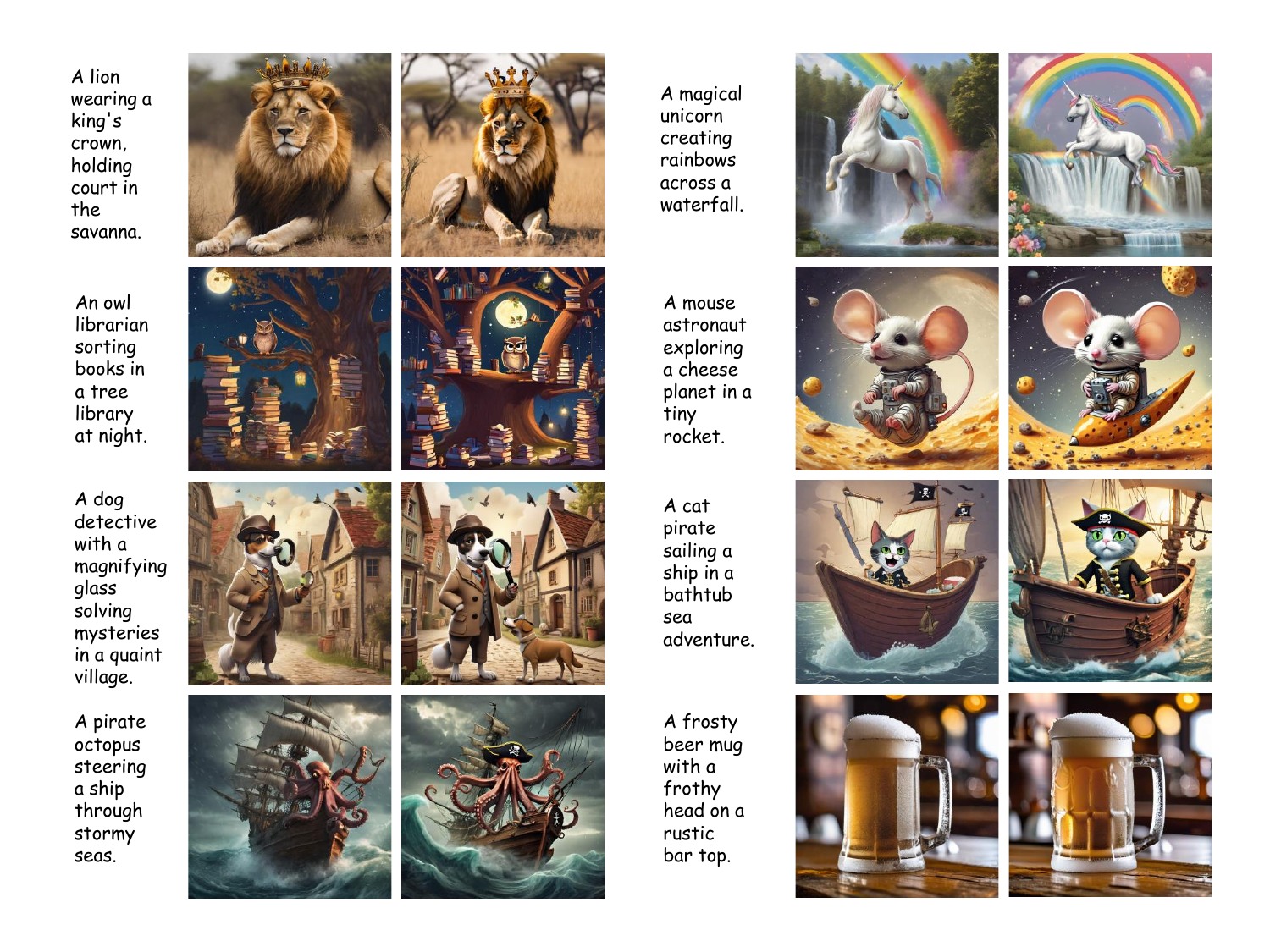}
    \caption{More examples generated by VersaT2I. Given the text prompt and fixed seed, we generate images with SDXL\cite{podell2023sdxl} and VersaT2I. Images generated by VersaT2I exhibit high quality.}
    \label{fig:s2}
\end{figure}

%% file: main.bbl
\begin{thebibliography}{10}
\providecommand{\url}[1]{\texttt{#1}}
\providecommand{\urlprefix}{URL }
\providecommand{\doi}[1]{https://doi.org/#1}

\bibitem{achiam2023gpt}
Achiam, J., Adler, S., Agarwal, S., Ahmad, L., Akkaya, I., Aleman, F.L., Almeida, D., Altenschmidt, J., Altman, S., Anadkat, S., et~al.: Gpt-4 technical report. arXiv preprint arXiv:2303.08774  (2023)

\bibitem{black2023training}
Black, K., Janner, M., Du, Y., Kostrikov, I., Levine, S.: Training diffusion models with reinforcement learning. arXiv preprint arXiv:2305.13301  (2023)

\bibitem{cao2023image}
Cao, S., Chai, W., Hao, S., Wang, G.: Image reference-guided fashion design with structure-aware transfer by diffusion models. In: Proceedings of the IEEE/CVF Conference on Computer Vision and Pattern Recognition. pp. 3524--3528 (2023)

\bibitem{cao2023difffashion}
Cao, S., Chai, W., Hao, S., Zhang, Y., Chen, H., Wang, G.: Difffashion: Reference-based fashion design with structure-aware transfer by diffusion models. arXiv preprint arXiv:2302.06826  (2023)

\bibitem{chai2023stablevideo}
Chai, W., Guo, X., Wang, G., Lu, Y.: Stablevideo: Text-driven consistency-aware diffusion video editing. In: Proceedings of the IEEE/CVF International Conference on Computer Vision. pp. 23040--23050 (2023)

\bibitem{chen2023sparse}
Chen, S., Ye, T., Bai, J., Chen, E., Shi, J., Zhu, L.: Sparse sampling transformer with uncertainty-driven ranking for unified removal of raindrops and rain streaks. arXiv preprint arXiv:2308.14153  (2023)

\bibitem{chen2022snowformer}
Chen, S., Ye, T., Liu, Y., Chen, E., Shi, J., Zhou, J.: Snowformer: Scale-aware transformer via context interaction for single image desnowing. arXiv preprint arXiv:2208.09703  (2022)

\bibitem{chen2022msp}
Chen, S., Ye, T., Liu, Y., Liao, T., Ye, Y., Chen, E.: Msp-former: Multi-scale projection transformer for single image desnowing. arXiv preprint arXiv:2207.05621  (2022)

\bibitem{chen2023dehrformer}
Chen, S., Ye, T., Shi, J., Liu, Y., Jiang, J., Chen, E., Chen, P.: Dehrformer: Real-time transformer for depth estimation and haze removal from varicolored haze scenes. In: ICASSP 2023-2023 IEEE International Conference on Acoustics, Speech and Signal Processing (ICASSP). pp.~1--5. IEEE (2023)

\bibitem{chen2023uncertainty}
Chen, S., Ye, T., Xue, C., Chen, H., Liu, Y., Chen, E., Zhu, L.: Uncertainty-driven dynamic degradation perceiving and background modeling for efficient single image desnowing. In: Proceedings of the 31st ACM International Conference on Multimedia. pp. 4269--4280 (2023)

\bibitem{cho2023davidsonian}
Cho, J., Hu, Y., Garg, R., Anderson, P., Krishna, R., Baldridge, J., Bansal, M., Pont-Tuset, J., Wang, S.: Davidsonian scene graph: Improving reliability in fine-grained evaluation for text-image generation. arXiv preprint arXiv:2310.18235  (2023)

\bibitem{deng2009imagenet}
Deng, J., Dong, W., Socher, R., Li, L.J., Li, K., Fei-Fei, L.: Imagenet: A large-scale hierarchical image database. In: 2009 IEEE conference on computer vision and pattern recognition. pp. 248--255. Ieee (2009)

\bibitem{deng2023citygen}
Deng, J., Chai, W., Guo, J., Huang, Q., Hu, W., Hwang, J.N., Wang, G.: Citygen: Infinite and controllable 3d city layout generation. arXiv preprint arXiv:2312.01508  (2023)

\bibitem{fan2023dpok}
Fan, Y., Watkins, O., Du, Y., Liu, H., Ryu, M., Boutilier, C., Abbeel, P., Ghavamzadeh, M., Lee, K., Lee, K.: Dpok: Reinforcement learning for fine-tuning text-to-image diffusion models. arXiv preprint arXiv:2305.16381  (2023)

\bibitem{hessel2021clipscore}
Hessel, J., Holtzman, A., Forbes, M., Bras, R.L., Choi, Y.: Clipscore: A reference-free evaluation metric for image captioning. arXiv preprint arXiv:2104.08718  (2021)

\bibitem{heusel2017gans}
Heusel, M., Ramsauer, H., Unterthiner, T., Nessler, B., Hochreiter, S.: Gans trained by a two time-scale update rule converge to a local nash equilibrium. Advances in neural information processing systems  \textbf{30} (2017)

\bibitem{ho2020denoising}
Ho, J., Jain, A., Abbeel, P.: Denoising diffusion probabilistic models. Advances in neural information processing systems  \textbf{33},  6840--6851 (2020)

\bibitem{hu2021lora}
Hu, E.J., Shen, Y., Wallis, P., Allen-Zhu, Z., Li, Y., Wang, S., Wang, L., Chen, W.: Lora: Low-rank adaptation of large language models. arXiv preprint arXiv:2106.09685  (2021)

\bibitem{hu2023tifa}
Hu, Y., Liu, B., Kasai, J., Wang, Y., Ostendorf, M., Krishna, R., Smith, N.A.: Tifa: Accurate and interpretable text-to-image faithfulness evaluation with question answering. arXiv preprint arXiv:2303.11897  (2023)

\bibitem{huang2023t2i}
Huang, K., Sun, K., Xie, E., Li, Z., Liu, X.: T2i-compbench: A comprehensive benchmark for open-world compositional text-to-image generation. In: Thirty-seventh Conference on Neural Information Processing Systems Datasets and Benchmarks Track (2023)

\bibitem{huang2023reversion}
Huang, Z., Wu, T., Jiang, Y., Chan, K.C., Liu, Z.: Reversion: Diffusion-based relation inversion from images. arXiv preprint arXiv:2303.13495  (2023)

\bibitem{jiang2023five}
Jiang, J., Ye, T., Bai, J., Chen, S., Chai, W., Jun, S., Liu, Y., Chen, E.: Five $a^+$ network: You only need 9k parameters for underwater image enhancement. arXiv preprint arXiv:2305.08824  (2023)

\bibitem{jin2023perspective}
Jin, L., Zhang, J., Hold-Geoffroy, Y., Wang, O., Blackburn-Matzen, K., Sticha, M., Fouhey, D.F.: Perspective fields for single image camera calibration. In: Proceedings of the IEEE/CVF Conference on Computer Vision and Pattern Recognition. pp. 17307--17316 (2023)

\bibitem{karthik2023if}
Karthik, S., Roth, K., Mancini, M., Akata, Z.: If at first you don't succeed, try, try again: Faithful diffusion-based text-to-image generation by selection. arXiv preprint arXiv:2305.13308  (2023)

\bibitem{ku2023viescore}
Ku, M., Jiang, D., Wei, C., Yue, X., Chen, W.: Viescore: Towards explainable metrics for conditional image synthesis evaluation. arXiv preprint arXiv:2312.14867  (2023)

\bibitem{lee2024parrot}
Lee, S.H., Li, Y., Ke, J., Yoo, I., Zhang, H., Yu, J., Wang, Q., Deng, F., Entis, G., He, J., et~al.: Parrot: Pareto-optimal multi-reward reinforcement learning framework for text-to-image generation. arXiv preprint arXiv:2401.05675  (2024)

\bibitem{lee2023holistic}
Lee, T., Yasunaga, M., Meng, C., Mai, Y., Park, J.S., Gupta, A., Zhang, Y., Narayanan, D., Teufel, H.B., Bellagente, M., et~al.: Holistic evaluation of text-to-image models. In: Thirty-seventh Conference on Neural Information Processing Systems Datasets and Benchmarks Track (2023)

\bibitem{li2022mplug}
Li, C., Xu, H., Tian, J., Wang, W., Yan, M., Bi, B., Ye, J., Chen, H., Xu, G., Cao, Z., et~al.: mplug: Effective and efficient vision-language learning by cross-modal skip-connections. arXiv preprint arXiv:2205.12005  (2022)

\bibitem{liang2023rich}
Liang, Y., He, J., Li, G., Li, P., Klimovskiy, A., Carolan, N., Sun, J., Pont-Tuset, J., Young, S., Yang, F., et~al.: Rich human feedback for text-to-image generation. arXiv preprint arXiv:2312.10240  (2023)

\bibitem{lin2014microsoft}
Lin, T.Y., Maire, M., Belongie, S., Hays, J., Perona, P., Ramanan, D., Doll{\'a}r, P., Zitnick, C.L.: Microsoft coco: Common objects in context. In: Computer Vision--ECCV 2014: 13th European Conference, Zurich, Switzerland, September 6-12, 2014, Proceedings, Part V 13. pp. 740--755. Springer (2014)

\bibitem{liu2023nighthazeformer}
Liu, Y., Yan, Z., Chen, S., Ye, T., Ren, W., Chen, E.: Nighthazeformer: Single nighttime haze removal using prior query transformer. arXiv preprint arXiv:2305.09533  (2023)

\bibitem{liu2022nighttime}
Liu, Y., Yan, Z., Wu, A., Ye, T., Li, Y.: Nighttime image dehazing based on variational decomposition model. In: Proceedings of the IEEE/CVF conference on computer vision and pattern recognition. pp. 640--649 (2022)

\bibitem{lu2023llmscore}
Lu, Y., Yang, X., Li, X., Wang, X.E., Wang, W.Y.: Llmscore: Unveiling the power of large language models in text-to-image synthesis evaluation. arXiv preprint arXiv:2305.11116  (2023)

\bibitem{nilsback2008automated}
Nilsback, M.E., Zisserman, A.: Automated flower classification over a large number of classes. In: 2008 Sixth Indian conference on computer vision, graphics \& image processing. pp. 722--729. IEEE (2008)

\bibitem{ouyang2023chasing}
Ouyang, Y., Chai, W., Ye, J., Tao, D., Zhan, Y., Wang, G.: Chasing consistency in text-to-3d generation from a single image. arXiv preprint arXiv:2309.03599  (2023)

\bibitem{park2021benchmark}
Park, D.H., Azadi, S., Liu, X., Darrell, T., Rohrbach, A.: Benchmark for compositional text-to-image synthesis. In: Thirty-fifth Conference on Neural Information Processing Systems Datasets and Benchmarks Track (Round 1) (2021)

\bibitem{podell2023sdxl}
Podell, D., English, Z., Lacey, K., Blattmann, A., Dockhorn, T., M{\"u}ller, J., Penna, J., Rombach, R.: Sdxl: Improving latent diffusion models for high-resolution image synthesis. arXiv preprint arXiv:2307.01952  (2023)

\bibitem{rombach2022high}
Rombach, R., Blattmann, A., Lorenz, D., Esser, P., Ommer, B.: High-resolution image synthesis with latent diffusion models. In: Proceedings of the IEEE/CVF conference on computer vision and pattern recognition. pp. 10684--10695 (2022)

\bibitem{ruiz2023dreambooth}
Ruiz, N., Li, Y., Jampani, V., Pritch, Y., Rubinstein, M., Aberman, K.: Dreambooth: Fine tuning text-to-image diffusion models for subject-driven generation. In: Proceedings of the IEEE/CVF Conference on Computer Vision and Pattern Recognition. pp. 22500--22510 (2023)

\bibitem{saharia2022photorealistic}
Saharia, C., Chan, W., Saxena, S., Li, L., Whang, J., Denton, E.L., Ghasemipour, K., Gontijo~Lopes, R., Karagol~Ayan, B., Salimans, T., et~al.: Photorealistic text-to-image diffusion models with deep language understanding. Advances in Neural Information Processing Systems  \textbf{35},  36479--36494 (2022)

\bibitem{salimans2016improved}
Salimans, T., Goodfellow, I., Zaremba, W., Cheung, V., Radford, A., Chen, X.: Improved techniques for training gans. Advances in neural information processing systems  \textbf{29} (2016)

\bibitem{sarkar2023shadows}
Sarkar, A., Mai, H., Mahapatra, A., Lazebnik, S., Forsyth, D.A., Bhattad, A.: Shadows don't lie and lines can't bend! generative models don't know projective geometry... for now. arXiv preprint arXiv:2311.17138  (2023)

\bibitem{shazeer2017outrageously}
Shazeer, N., Mirhoseini, A., Maziarz, K., Davis, A., Le, Q., Hinton, G., Dean, J.: Outrageously large neural networks: The sparsely-gated mixture-of-experts layer. arXiv preprint arXiv:1701.06538  (2017)

\bibitem{sohn2023styledrop}
Sohn, K., Ruiz, N., Lee, K., Chin, D.C., Blok, I., Chang, H., Barber, J., Jiang, L., Entis, G., Li, Y., et~al.: Styledrop: Text-to-image generation in any style. arXiv preprint arXiv:2306.00983  (2023)

\bibitem{sun2023dreamsync}
Sun, J., Fu, D., Hu, Y., Wang, S., Rassin, R., Juan, D.C., Alon, D., Herrmann, C., van Steenkiste, S., Krishna, R., et~al.: Dreamsync: Aligning text-to-image generation with image understanding feedback. arXiv preprint arXiv:2311.17946  (2023)

\bibitem{wah2011caltech}
Wah, C., Branson, S., Welinder, P., Perona, P., Belongie, S.: The caltech-ucsd birds-200-2011 dataset  (2011)

\bibitem{wallace2023diffusion}
Wallace, B., Dang, M., Rafailov, R., Zhou, L., Lou, A., Purushwalkam, S., Ermon, S., Xiong, C., Joty, S., Naik, N.: Diffusion model alignment using direct preference optimization. arXiv preprint arXiv:2311.12908  (2023)

\bibitem{wu2023q-instruct}
Wu, H., Zhang, Z., Zhang, E., Chen, C., Liao, L., Wang, A., Xu, K., Li, C., Hou, J., Zhai, G., et~al.: Q-instruct: Improving low-level visual abilities for multi-modality foundation models. arXiv preprint arXiv:2311.06783  (2023)

\bibitem{wu2023q}
Wu, H., Zhang, Z., Zhang, W., Chen, C., Liao, L., Li, C., Gao, Y., Wang, A., Zhang, E., Sun, W., et~al.: Q-align: Teaching lmms for visual scoring via discrete text-defined levels. arXiv preprint arXiv:2312.17090  (2023)

\bibitem{wu2023human}
Wu, X., Hao, Y., Sun, K., Chen, Y., Zhu, F., Zhao, R., Li, H.: Human preference score v2: A solid benchmark for evaluating human preferences of text-to-image synthesis. arXiv preprint arXiv:2306.09341  (2023)

\bibitem{xu2023imagereward}
Xu, J., Liu, X., Wu, Y., Tong, Y., Li, Q., Ding, M., Tang, J., Dong, Y.: Imagereward: Learning and evaluating human preferences for text-to-image generation. arXiv preprint arXiv:2304.05977  (2023)

\bibitem{ye2023adverse}
Ye, T., Chen, S., Bai, J., Shi, J., Xue, C., Jiang, J., Yin, J., Chen, E., Liu, Y.: Adverse weather removal with codebook priors. In: Proceedings of the IEEE/CVF International Conference on Computer Vision. pp. 12653--12664 (2023)

\bibitem{ye2022towards}
Ye, T., Chen, S., Liu, Y., Ye, Y., Bai, J., Chen, E.: Towards real-time high-definition image snow removal: Efficient pyramid network with asymmetrical encoder-decoder architecture. In: Proceedings of the Asian Conference on Computer Vision. pp. 366--381 (2022)

\bibitem{ye2022underwater}
Ye, T., Chen, S., Liu, Y., Ye, Y., Chen, E., Li, Y.: Underwater light field retention: Neural rendering for underwater imaging. In: Proceedings of the IEEE/CVF Conference on Computer Vision and Pattern Recognition. pp. 488--497 (2022)

\bibitem{ye2021perceiving}
Ye, T., Zhang, Y., Jiang, M., Chen, L., Liu, Y., Chen, S., Chen, E.: Perceiving and modeling density for image dehazing. In: European Conference on Computer Vision. pp. 130--145. Springer (2022)

\bibitem{zhang2024composing}
Zhang, J., Liu, J., He, J., et~al.: Composing parameter-efficient modules with arithmetic operation. Advances in Neural Information Processing Systems  \textbf{36} (2024)

\bibitem{zhang2023adding}
Zhang, L., Rao, A., Agrawala, M.: Adding conditional control to text-to-image diffusion models. In: Proceedings of the IEEE/CVF International Conference on Computer Vision. pp. 3836--3847 (2023)

\bibitem{zhang2018unreasonable}
Zhang, R., Isola, P., Efros, A.A., Shechtman, E., Wang, O.: The unreasonable effectiveness of deep features as a perceptual metric. In: Proceedings of the IEEE conference on computer vision and pattern recognition. pp. 586--595 (2018)

\bibitem{zou2023vqcnir}
Zou, W., Gao, H., Ye, T., Chen, L., Yang, W., Huang, S., Chen, H., Chen, S.: Vqcnir: Clearer night image restoration with vector-quantized codebook. arXiv preprint arXiv:2312.08606  (2023)

\bibitem{zou2022self}
Zou, W., Ye, T., Zheng, W., Zhang, Y., Chen, L., Wu, Y.: Self-calibrated efficient transformer for lightweight super-resolution. In: Proceedings of the IEEE/CVF Conference on Computer Vision and Pattern Recognition. pp. 930--939 (2022)

\end{thebibliography}
